\documentclass{article}
\usepackage{spconf,amsmath,graphicx}
 \usepackage{url,color}
\usepackage[pdftex=true,hyperindex=true,colorlinks=true]{hyperref}
\def\x{{\mathbf x}}

\title{Max-Min convolutional neural networks for image classification}
\name{Michael Blot, Matthieu Cord, Nicolas Thome\thanks{Thanks to DGA for funding.}}
\address{Sorbonne Universit\'es, UPMC Univ Paris 06, CNRS, LIP6 UMR 7606, 4 place Jussieu 75005 Paris}
\begin{document}
%
\maketitle
\begin{abstract}
Convolutional neural networks (CNN) are widely used in computer vision, especially in image classification. 
However, the way in which information and invariance properties are encoded through in deep CNN architectures is still an open question.

 In this paper, we propose to modify the standard convolutional block of CNN in order to transfer more information layer after layer while keeping some invariance within the network. Our main idea is to exploit both positive and negative high scores obtained in the convolution maps. This behavior is obtained by modifying the traditional activation function step before pooling. We are doubling the maps with specific activations functions, called MaxMin strategy, in order to achieve our pipeline. Extensive experiments on two classical datasets, MNIST and CIFAR-10, show that our deep MaxMin convolutional net outperforms standard CNN.
\end{abstract}
\section{Introduction}
\label{sec:intro}

Computer vision has for a long time great interaction with artificial intelligence and machine learning. One of the main examples studied in this article is image classification.  

For a while, state of the art algorithms for image classification were based on bag of words (BoW) models \cite{bow1} \cite{bow2}.
~Those algorithms build a visual dictionary from local image descriptors. For any image, local detection values of each word are pooled together to represent the final image feature, which is used as input for a SVM \cite{svm} classifier.
Several attempts have been made to improve the coding or pooling steps~\cite{DBLP:conf/cvpr/PerronninD07,Goh_NIPS13,GohTNNLS14,avilaCVIU2013}.

Recently, a major breakthrough has been revealed with convolutional neural networks (CNN)~\cite{alexnet} beating all others models with a huge gap on the ILSVR2012 competition \cite{lsvrc2012}. 
An appealing feature of deep learning is the ability to learn representations from raw pixels. 
Today CNN introduced for the first time in \cite{convnet1} and popularized by \cite{convnet2} are widely 
used is computer vision and stand as the state of the art in image classification on many popular datasets. 

The CNN architecture can be described as a succession of convolutional blocks followed by some fully connected layers. 
Classically, one convolution block successively applies linear filters, non-linear activation functions (like $ReLU$), and local pooling operations \cite{alexnet}.

In BoW, the pooling phase enables to manage many invariances but it is also responsible for a loss of spatial information that make the learning difficult. In deep convolutional architectures, activation functions and pooling may be also questioned. For instance, using a $ReLU$ activation function, all negative information is removed from the convolutional map considered. On the contrary, we propose to modify the CNN layers by introducing a new block to preserve this information from strong negative detections. We assume that keeping both positive and negative evidences for each filter may be of interest for the whole architecture. Inspired by \cite{bossanova, avilaCVIU2013}, we propose a new pipeline doubling the output map for each linear filter in order to independently process these maps with dedicated activation functions. The resulting deep CNN architecture, called  MaxMin CNN, is evaluated against simple CNN on two benchmarks.

\section{RELATED WORKS AND MOTIVATIONS}
\label{sec:format}
The standard deep CNN introduced in \cite{alexnet} is composed with 5 convolutional blocks followed by 3 fully connected layers with a Softmax function at the end. 
Each convolutional block has a convolutional filter layer applying a number of filters to the input and concatenates the resulting maps of convolution. This output passes through an activation function that is responsible for increasing the representational power of the network. In image classification the mostly used function is $ReLU$ \cite{alexnet}. Then frequently comes a pooling layer  \cite{theoretical}. Aggregating information in a local neighborhood, this step ensures a CNN invariance to small translations. Optionally, others kind of layers such as local response normalisation may be used \cite{alexnet}. 

Since \cite{alexnet}, many CNN architecture improvements have been proposed to boost CNN performances. For instance \cite{vgg} and \cite{googlenet} improved greatly the performances of CNN on ImageNet by using a deeper and optimized architecture. 

Concerning the pooling function \cite{theoretical} questions about the nature of the information transmitted and proposed parametric functions that include max and average. Varying the parameters they tried to optimise the pooling function but obtained no better results that average or max pooling showing that it is difficult to improve the pooling function itself. They also made a theoretical analysis explaining the conditions on the input distribution involving max pooling working better than average pooling. Still on pooling  \cite{fractional} proposes a method selecting random smaller pooling windows enabling to filter less information that will be exploited with more convolutional layers.  

Regarding $ReLU$, it is a non-linear function that sets to zero all negative values after the convolutional layer. This filtering is supposed to facilitate the exploitation of discriminative information by de-noising filter detections. \cite{prelu} introduced the $PReLU$ as an extension of $ReLU$. It multiplies negative values with a learnable coefficient instead of setting them to zero. This method enables to filter less information at activation layer while keeping the non linear property of $ReLU$ although at the cost of additional learnable parameters. 

We explore in this paper a different strategy to improve convolutional block at the $ReLU$ step. As mentioned, $ReLU$ and pooling filter a lot of information from their input. 
We modify the convolutional block in order to keep more information after the $ReLU$. Indeed, the information from strong negative detection is important and is totally left out after the $ReLU$. Keeping and exploiting this special information is the goal of our method. 
Inspired by \cite{bossanova, avilaCVIU2013} which modifies the pooling by using a histogram of pattern detection instead of just keeping the maximum, we propose to keep high positive and high negative values obtained after filtering in a double $ReLU$ scheme (MaxMin) that we detail in the following. Similar researches have been conducted concurrently in \cite{crelu} with a new activation function called CReLU. 

\section{MaxMin CNN}
\label{sec:pagestyle}

\subsection{MaxMin Net Architecture}
\label{ssec:subhead}
With the $ReLU$ strong negative detections give the same information as weak negative detections. Unfortunately strong negative detections can be discriminant. 
Similarly than with geometrical shapes the semantic of a pattern can be invariant to the negative transformation. 
Thus it seems natural to presume that for many patterns that have important discriminant power the network learns the filters that detects the patterns as well as their opposites. 
Our method aims at transmitting directly the negative detections from a filter in order to prevent the network to learn the opposite filter. Indeed for a filter $h$ the negative filter $h^{-}=-h$ verifies for all input $\x$, $\x*h^{-} = \x *(-h) = - (\x*h)$. Then if the pattern filtered by $h^{-}$ is strongly detected on $\x$ then $\x*h^{-}$ will be high and positive while $\x*h$ will be symmetrically low and negative. Thus conserving the information from the convolution with $h$ either the result is positive or negative will give the information about $h^{-}$ and enables to use less filter in the architecture.

\begin{figure}[htb]
\centering
   \includegraphics[width=9cm, height=5cm]{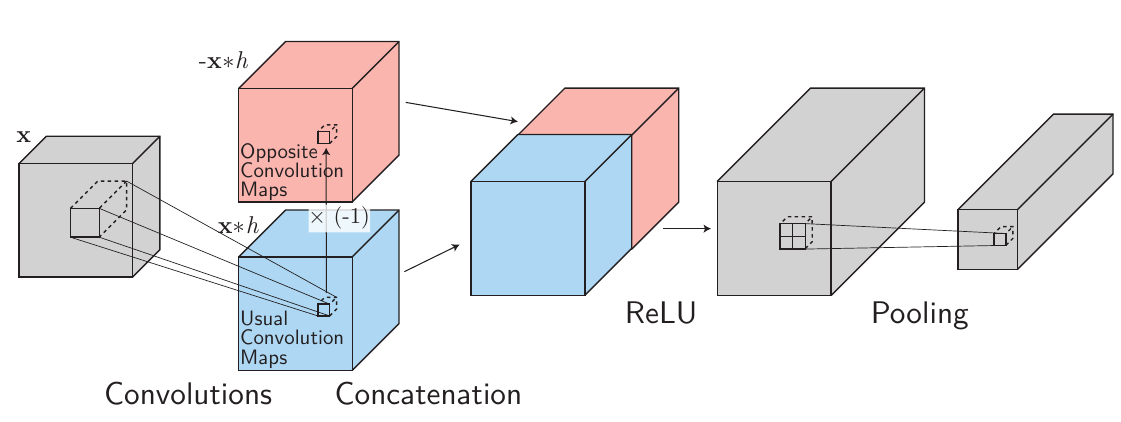}
\caption{MaxMin scheme. After the convolutional layer, the resulting maps (in blue) are duplicated negatively (in red). Both stacked maps are concatenated to get the MaxMin output that will pass through the ReLU before pooling.}
\label{scheme1}
\end{figure}

To implement our strategy, we duplicate the convolutional filter maps (represented is blue in Fig. \ref{scheme1}) and multiply the copy by $\times -1$ resulting in the negative version of the detections (in red in Fig. \ref{scheme1}). 
We then concatenate the original maps and their negative copy as shown on Fig. \ref{scheme1}. This operation increases the depth of the convolutional layer's output by two compared to classical CNN. We then apply the $ReLU$ normally to the concatenated output and optionally process a pooling operation. This way negative values are not filtered and can be exploited by following layers.

\vspace*{-0.6cm}
\subsection{Relation with max pooling}
\label{sssec:subsubhead}
After the $ReLU$ is frequently applied max pooling map by map. Let first remark that $ReLU \circ max = max \circ ReLU$. It is possible to commute the two layers without changing the block's output. For our method we notice the fact that 
\begin{align*}
\Big(max (ReLU(X)),\ max (ReLU(-X))\Big) &= \\ 
 \Big(ReLU (max(X)),&\ ReLU (-min(X))\Big)
\end{align*}
where $X$ is a vector. Thus our method can be interpreted as simply adding an additional information at pooling function with a bi-dimensional output when applied before the $ReLU$. This additional information is the minimum detection on the window taken negatively. 

\begin{figure*}[htb]
\centering
   \includegraphics[width=15cm]{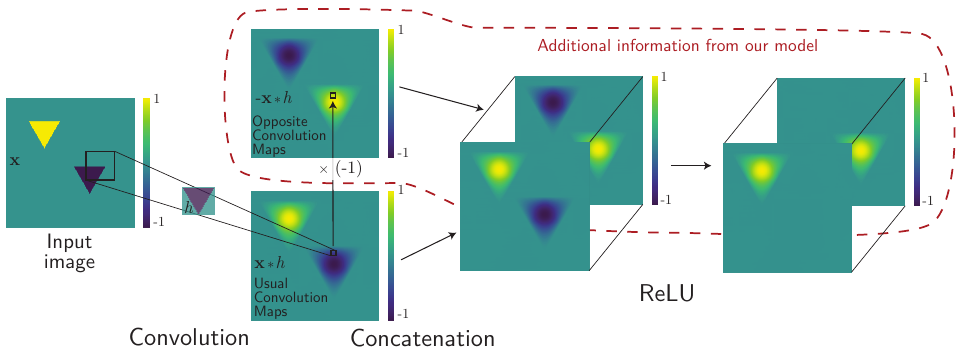}
\caption{MaxMin CNN block for one input image with specific patterns. When considering input with 2 specific triangles and a convolution with the filter $h$, we obtain a map with 2 strong positive and negative answers. Thanks to our MaxMin, we get an additional information (red dotted line) that makes possible to keep both extreme information after ReLU in a joint double map.}
\label{scheme2}
\end{figure*}

\subsection{Discussion}
Here we discuss the interest of the method in term of information modeling the generalization improvement.
\vspace*{-0.3cm}\paragraph*{Modeling:}
As the size of the output of the convolutional layer is doubled within the convolutional block it is necessary that the size of the filter of the following block is doubled too resulting on additional parameters on following filters. It worth mentioning that if those additional parameters are all set to zero the network is equivalent to a simple CNN and our method does not reduce CNN representational power. 

At the convolutional layer MaxMin network enforces the transmission of the detection from both the filters and their negative versions. The learning of the negative version of all filters is then not necessary. This is very useful for pattern that are invariant to negative transformations. Fig.~ \ref{scheme2} shows that different versions of a same shape (yellow and blue triangles) can be transmitted with MaxMin network with only 1 filter, whereas 2 filters are necessary for standard CNN. Thus we can reduce the number of filters and keep a reasonable amount of parameters.
\vspace*{-0.6cm}\paragraph*{Generalization:}
On a spatial window, the maximum and the minimum pooling outputs are rarely simultaneously positively high and negatively low. Therefore, $ReLU$ filters very often one of the two values (min or max). This ensures a sparse activation of the network's neurons. This property is known to enhance the generalization performance of neural networks in computer vision as studied in \cite{sparse} and \cite{dropout}. \\
Moreover, we claim that our method is able to learn more efficiently the convolutional filters than standard CNN. MaxMin method enforces discriminant patterns to be learnt by back-propagating the error from both positive detections and negative ones. MaxMin networks learn each pattern both from its positive and negative occurrences in the dataset. They learn the pattern filters more accurately and faster than a classical CNN that would learn positive and negative filters independently. 

\section{Experiments}
\label{sec:typestyle}
We test our method on two well-known datasets, MNIST and CIFAR-10, and we compare our results with classical CNN. 

\paragraph*{Learning protocol:}
All the models and the learning framework are implemented in Lua using Torch 7 library in \cite{torch} \footnote{Code will be released on Gitub after publication.}. For the training, we use the same gradient descent as in \cite{alexnet} with 0.9 of momentum and fixed weight decay ($10^{-3}$ for MNIST and $10^{-4}$ for CIFAR). We use an equal learning rate for all layers that is being manually reduced when validation error stops decrease. All weights of convolutional filters are initialized from a zero-mean Gaussian distribution with standard deviation 0.01. The number of training epochs depends on the network but is between 30 and 120 for CIFAR-10 and around 250 for MNIST. Top-1 accuracy on test set is computed to evaluate performances.

\subsection{MNIST RESULTS}
MNIST is a 60,000 images dataset representing 9 handwritten digits. Images are $32 \times 32$ pixels in one channel. See \cite{multicolumn} for more information. We use the same protocol as \cite{multicolumn}: 50,000 images for training and the 10,000 remaining images for testing. To compare our method,  we use a LeNet like network with pooling layers. It is composed of three convolutional layers with $ReLU$ activations all followed by max pooling and a local contrast normalisation layers. The filters are of size $5 \times 5$ with a stride of 1. There is 64 of them at each convolutional layers. The pooling windows are of size $3\times 3$ for a stride of 2. Those layers are followed by one fully connected layers before a softmax. 

For the MaxMin network setting, we simply include a MaxMin layer before the $ReLU$ activation function. We keep the same number of filters paying attention to double their size when needed. 

The baseline network obtains a score of 99.34\% accuracy. With our MaxMin network, we obtain a score of 99.39\% accuracy.
Our strategy has 31 errors on the validation set against 36 errors for the baseline CNN. It corresponds to an relative improvement of 13.9\%.    Note that the performance scores on this dataset are very high, and any small performance gain is difficult to obtain.                
                                                                                  
\subsection{CIFAR-10 RESULTS}
CIFAR-10 dataset consists of 60,000 $32 \times 32$ color images (50,000 training images and 10,000 test images) representing 10 objects class like airplanes, trucks or birds. Images vary greatly within each class. See \cite{multicolumn} for more information. The baseline network is the one implemented in \cite{matconvnet} where we have replaced the average pooling layers with max pooling. The network has three convolutional layers with $ReLU$ activation, all followed by max pooling. All filters have size $5 \times 5$ and stride 1 with 32 filters at the first two layers and 64 at the last one. The pooling windows have size $3\times 3$ and stride 2 at all layers. The last max pooling layer is followed by two fully connected layers before a softmax. 

The accuracy score obtained using this simple CNN is 74.44\% on the test set.
When we train an equivalent MaxMin network, we reach an accuracy of 78.62\%, showing an improvement of more than 4\% of the results. 
\paragraph*{Robustness to parameter number:}
\begin{table}
\begin{center}
\begin{tabular}{|l|c|r|}
   \hline
   \# Parameters & MaxMin-CNN & Simple-CNN \\
   \hline
	  $ \approx 30K$ & 73.81 & 69.98 \\
   \hline
      $ \approx 15K$ & 78.13 & 74.44 \\
   \hline
      $ \approx 60K$ & 80.03 & 77.01 \\
   \hline
      $ \approx 2M$  & 81.68 & 78.11 \\
   \hline
      $ \approx 5M$  & 82.07 & 79.48 \\
   \hline   
      $ \approx 15M$ & 82.69 & 80.13 \\
   \hline 
      $ \approx 45M$ & 82.98 & 80.41 \\
   \hline 
\end{tabular}
\caption{CIFAR-10 results for simple CNN and our MaxMin CNN. Accuracy scores are reported for different deep architectures parametrized by their number of parameters. MaxMin network has systematically better performances than simple CNN for equal complexity (\# Parameters).}
\label{resultscifar1}
\end{center}
\end{table}
As mentioned earlier, our MaxMin layer imposes to double up the size of following filters compared to classical CNN. At fixed number of filters, this  implies to increase the number of parameters when including MaxMin layers to CNN. As the number of parameters directly impacts the performances of a CNN, we now compare both methods with constant number of parameters. We  use the previous network and modify the number of filters on different convolutional layers in order to vary the total number of parameters. For the MaxMin network, we adapt the number of filters to get comparable amount of parameters as CNN networks. As we intend to compare the quality of the exploitation of the filters by CNN and MaxMin, we pay attention to get the same amount of neurons in the fully connected layers. Only the number of convolutional filters varies from a method to another.
We report in table \ref{resultscifar1} the accuracies obtained for different numbers of parameters for both simple CNN and MaxMin networks.
We observe that the MaxMin network systematically outperforms the CNN whatever the number of parameters is with a best performance of 82.98\%. This shows the robustness of our method to different network parametrization and demonstrates the ability of MaxMin networks to exploit filter information.
\paragraph*{Boosting performances:}
Current deep models on CIFAR-10 use several kinds of learning tricks such as data augmentation, image preprocessing, dropout, to improve final classification performances. 

Our MaxMin deep convolutional model may benefit of the same tricks. As the learning becomes quite more complex and much more time consuming, we do perform only once these optimizations for our MaxMin architecture model on CIFAR-10. We thus apply some translations and flips for data augmentation, zca whitening referenced in \cite{zca} to preprocess the images, and dropout from \cite{dropout} to boost the learning. We use the same network described above with additional local contrast normalisation layer after each pooling. 

With this training of our network, we do obtain the accuracy of 90.03\% on the test set.  This score beats the results reported by \cite{alexnet} that uses a deeper architecture ($89\%$) and reported by \cite{multicolumn} that uses 8 networks for prediction (88.79\%). On the contrary, our MaxMin scores are obtained with only one network. Our method could also benefit from the use of several nets for prediction in order to reach the performances of  recent very deep architectures as \cite{cif1} or \cite{cif2} with an accuracy of 92.40\%. 

\section{CONCLUSION}
\label{sec:majhead}
In this paper, we  present our new deep CNN architecture, MaxMin-CNN, to better encode both positive and negative filter detections in the net. MaxMin strategy aims at preserving and propagating significant negative detection values through the net. This difference with standard CNN enables us  to preserve and transfer more information through the network. 

We  evaluate and compare our strategy to classical deep CNNs on two benchmarks CIFAR-10 and MNIST. 
Results demonstrate that our MaxMin networks perform better than CNNs, whatever the configuration considered.

Finally, our MaxMin strategy reaches very good performances on CIFAR-10 outperforming several recent and much more complex deep architectures. The principle is very simple and many deep  networks could benefit of this MaxMin strategy with only minor adaptations.  




\newpage
\newpage

\bibliographystyle{IEEE}
\bibliography{Biblio}

\end{document}